\documentclass[twoside]{article}

\usepackage[preprint]{aistats2026}
\usepackage{amsmath,amssymb}
\usepackage{graphicx}
\usepackage{svg}
\usepackage{booktabs}
\usepackage{natbib}
\usepackage{url}
\renewcommand{\url}[1]{}
\newcommand{\urldate}[1]{}
\allowdisplaybreaks

\begin{document}
\AtBeginDocument{%
  \setlength\abovedisplayskip{0pt}
  \setlength\belowdisplayskip{0pt}}

%

%

\twocolumn[

\aistatstitle{Learning with Embedded Linear Equality Constraints via Variational Bayesian Inference}

\aistatsauthor{ Matthew Marsh \And Benoît Chachuat \And  Antonio del Rio Chanona }

\aistatsaddress{
Sargent Centre for Process Systems Engineering, 
Imperial College London
} ]

\begin{abstract}
Machine Learning is becoming more prevalent in science and engineering, but many approaches do not provide meaningful uncertainty estimates and predictions may also violate known physical knowledge. We propose a Bayesian framework to embed linear relationships across inputs and outputs into the learning process, whilst characterizing full predictive uncertainty over both the model parameters and the domain knowledge. We evaluated our method on learning the single particle battery model subject to voltage and energy balances, showing its ability to provide reduced credible intervals and constraint violations compared to standard Bayesian neural networks based on variational inference.
\end{abstract}

\section{Introduction}

In many scientific and engineering domains, predictive models must satisfy known physical constraints. These constraints are often linear equalities arising from mass and energy balances. However, modern neural networks trained purely on data, of limited quantity or quality, with no consideration of constraints may frequently violate such relations, leading to physically inconsistent or infeasible predictions.

Bayesian neural networks (BNNs) provide a principled framework for uncertainty quantification by placing distributions over model parameters \citep{nealBayesianLearningNeural1996, blundellWeightUncertaintyNeural}. Yet, standard BNN formulations do not explicitly enforce known constraints in the predictive distribution. While hard projection methods and penalty-based approaches exist \citep{raissiPhysicsinformedNeuralNetworks2019, amosOptNetDifferentiableOptimization2017, dontiDC3LearningMethod2021}, they typically treat constraints deterministically and do not account for overall uncertainty in constraint satisfaction.

To overcome these challenges, we propose a probabilistic framework for embedding linear equality constraints directly into BNNs, tractably evaluated using variational inference (VI). We show that when the predictive distribution is Gaussian and the constraints are linear, constraint enforcement reduces to closed-form Gaussian conditioning. This yields a posterior predictive distribution that satisfies constraints up to a defined tolerance level while preserving calibrated uncertainty.

Furthermore, we treat the constraint tolerance as a random variable and perform VI jointly over the network parameters and tolerance level. Alongside the constraint embedding, this allows the model to learn how strictly the constraints should be enforced from data, rather than assuming they hold exactly. The resulting framework integrates structured knowledge, uncertainty quantification, and tractable optimisation into a single differentiable training objective, providing a principled bridge between structured probabilistic modelling and modern deep learning.

\section{Related Work}

Neural networks are powerful approximators, however standard deterministic training yields only point predictions with no principled measure of confidence, and offers no mechanism to enforce known physical relationships. In science and engineering applications, both are important limitation: uncalibrated predictions may be overconfident, and physically inconsistent outputs undermine trust and decision-making. Bayesian inference and constraint enforcement address these shortcomings independently, but existing approaches rarely combine them. Whilst Bayesian methods quantify uncertainty without enforcing structure, constraint methods enforce structure without propagating uncertainty. 

\subsection{Bayesian Neural Networks}

Bayesian inference provides a natural framework for uncertainty quantification (UQ) by treating model parameters as random variables whose posterior distribution reflects the evidence provided by data. Exact inference is intractable for deep networks, motivating scalable approximations such as VI \citep{gravesPracticalVariationalInference2011, blundellWeightUncertaintyNeural} and Monte Carlo dropout \citep{galDropoutBayesianApproximation2016}. These methods provide predictive uncertainty estimates, though without considering constraint satisfaction.

\subsection{Constraining Neural Networks}

Physics-informed neural networks (PINNs) incorporate physical knowledge by penalising physical constraint residuals during training \citep{raissiPhysicsinformedNeuralNetworks2019}, and are amongst the most widely adopted example of \emph{soft} constraint enforcement. However, this approach does not guarantee that the constraints are enforced everywhere, neither during training nor inference. 

Hard constraints, on the other hand, augment model architectures to enforce constraints on all model outputs. Some approaches utilise projections onto feasible outputs \citep{chenPhysicsinformedNeuralNetworks2024}. OptNet \citep{amosOptNetDifferentiableOptimization2017} uses quadratic programs as differentiable layers, expanded in subsequent work to general differentiable convex optimisation layers \citep{agrawalDifferentiableConvexOptimization2019}. Subspace-based methods can also be used to reconstruct outputs consistent with constraints \citep{dontiDC3LearningMethod2021}. 

A common limitation shared by both soft and hard approaches is that constraint satisfaction is often treated deterministically, thereby omitting uncertainty propagation. Our work addresses this gap by incorporating constraints probabilistically via Gaussian conditioning \citep{hansenLearningPhysicalModels2023}, treating constraints as noisy linear observations, with joint inference yielding a framework that enforces physical structure while preserving uncertainty estimates.

\subsection{Post-Bayesian Inference and Structured Conditioning}

In classical Gaussian models, conditioning under linear transformations admits closed-form solutions \citep{pml2Book}. Our approach integrates linear constraint conditioning directly into the variational objective, with the posterior learned jointly with a probabilistic tolerance over the constraints. The method can be interpreted as a post-Bayesian extension: rather than performing Bayesian inference over unconstrained function classes, uncertainty is conditioned and reshaped by known linear relationships, embedding inductive bias directly into the model.

\section{Methodology}

We consider supervised learning over a dataset $\mathcal{D} = \{(\mathbf{x}_i, \mathbf{y}_i)\}_{i=1}^N$, with $\mathbf{x} \in \mathbb{R}^{n_x}$, $\mathbf{y} \in \mathbb{R}^{n_y}$. The neural network outputs a diagonal Gaussian, with mean $\boldsymbol{\mu}_P\in \mathbb{R}^{n_y}$ and variance $\boldsymbol{\sigma}^2_P \in \mathbb{R}^{n_y}$:
\begin{equation}
p(\mathbf{y} \mid \mathbf{x}, \boldsymbol{\theta})
=
\mathcal{N}\left(\boldsymbol{\mu}_P(\mathbf{x};\boldsymbol{\theta}),
\mathrm{diag}(\boldsymbol{\sigma}_P^2(\mathbf{x};\boldsymbol{\theta}))\right),
\end{equation}
with the neural network parameters denoted by $\boldsymbol{\theta}\in\mathbb{R}^{n_\theta}$.

\subsection{Embedding Constraints with Linear-Gaussian System}

We assume $m$ known linear equality relationships between input $(\mathbf{x})$ and predicted output $(\mathbf{y})$ variables. As our inputs are treated as fixed, we introduce a small tolerance term ($\boldsymbol{\varepsilon}$) to account for measurement noise. 

The constraint residual is given by:
\begin{equation}
\mathbf{z}
=
\mathbf{A}\mathbf{x} + \mathbf{B}\mathbf{y} - \mathbf{b}
+ \boldsymbol{\varepsilon},
\quad
\boldsymbol{\varepsilon} \sim \mathcal{N}(\mathbf{0}, \operatorname{diag}(\mathbf{r})),
\end{equation}
where $\mathbf{A} \in \mathbb{R}^{m \times n_x}, \mathbf{B} \in \mathbb{R}^{m \times n_y}, \mathbf{b} \in \mathbb{R}^m$, with $\mathbf{B}$ of full row rank, and $\mathbf{r} \in \mathbb{R}^m$ is the diagonal variance parameterizing the tolerance level.

Conditioning on the residual of the constraint, $\mathbf{z}=\mathbf{0}$ yields a Gaussian posterior predictive:
\begin{equation}\label{eq:cond_Gaussian}
p(\mathbf{y} \mid \mathbf{x}, \boldsymbol{\theta}, \mathbf{z}=\mathbf{0})
=
\mathcal{N}(\boldsymbol{\mu}_C, \operatorname{diag(}\boldsymbol{\sigma}^2_C)),
\end{equation}
with closed-form updated parameters given by:
\begin{align}
\mathbf{S} &= \mathbf{B} \operatorname{diag}(\boldsymbol{\sigma}_P^2) \mathbf{B}^{\top} + \boldsymbol{\varepsilon} \label{eq:schur} \\
\!\!\boldsymbol{\mu}_C &= \boldsymbol{\mu}_P + \left( \boldsymbol{\sigma}_P^2 \odot \mathbf{B}^{\top} \right) \mathbf{S}^{-1} \left( \mathbf{b} - \mathbf{A}\mathbf{x} - \mathbf{B}\boldsymbol{\mu}_P \right) \label{eq:mu_update} \\
\!\!\boldsymbol{\sigma}_C^2 &= \operatorname{diag}(\boldsymbol{\sigma}_P^2) - \left( \boldsymbol{\sigma}_P^2 \odot \mathbf{B}^{\top} \right) \mathbf{S}^{-1} \left( \mathbf{B} \odot (\boldsymbol{\sigma}_P^2)^{\top} \right). \label{eq:sigma_update}
\end{align}

This conditioning layer is fully differentiable and can be inserted into the network as a probabilistic projection operator. Importantly, although this update step induces cross correlation within the updated variance, we only consider the diagonal elements here to reduce number of learnable parameters. 

The method recovers exact hard constraint enforcement as $\mathbf{r} \to 0$, while reverting to unconstrained predictions when $\mathbf{r}$ is large. Thus, $\mathbf{r}$ controls the strength of constraint enforcement and it becomes a learnable quantity within the Bayesian framework.

\subsection{Bayesian Inference over Constrained Neural Networks}

\paragraph{Joint Variational Inference.}
Conventional BNNs perform inference over the sole model parameters, whereas we perform joint inference over both the parameters $\boldsymbol{\theta}$ and constraint tolerance $\mathbf{r}$ here. We aim to find a tractable mean-field variational approximation, $q(\boldsymbol{\theta}, \mathbf{r})$, to the true posterior, $p(\boldsymbol{\theta}, \mathbf{r} \mid \mathcal{D})$. We approximate both the priors and variational posterior jointly over the network parameters and constraint tolerances as independent:
\begin{align*}
p(\boldsymbol{\theta}, \mathbf{r}) = p(\boldsymbol{\theta})p(\mathbf{r}), && q(\boldsymbol{\theta}, \mathbf{r}) = q(\boldsymbol{\theta}) q(\mathbf{r}),
\end{align*}
noting that $\mathbf{r}$ must be positive as the prior variance on our constraint tolerance.

\paragraph{Modified ELBO.}
The predictive likelihood incorporates the conditioned Gaussian distribution in Eqn.~\eqref{eq:cond_Gaussian}. The variational objective is derived by minimizing the KL divergence between the variational and true posterior, yielding the negative ELBO:
\begin{align*}
q^*(\boldsymbol{\theta}, \mathbf{r})
&\in \arg\min_{q(\boldsymbol{\theta}, \mathbf{r})}
D_{\mathrm{KL}}\!\left(
q(\boldsymbol{\theta}, \mathbf{r})\,\|\, p(\boldsymbol{\theta}, \mathbf{r} \mid \mathcal{D})
\right)
\\
&=
\mathbb{E}_{q(\boldsymbol{\theta}, \mathbf{r})}
\!\left[
\log \frac{q(\boldsymbol{\theta}, \mathbf{r})}
{p(\mathcal{D}\mid\boldsymbol{\theta}, \mathbf{r})p(\boldsymbol{\theta}, \mathbf{r})}
\right]
\\
&=
\mathbb{E}_{q(\boldsymbol{\theta})q(\mathbf{r})}
[-\log p(\mathcal{D}\mid\boldsymbol{\theta}, \mathbf{r})]
\\
&\quad
+ D_{\mathrm{KL}}\!\left(q(\mathbf{r})\,\|\,p(\mathbf{r})\right)
+ D_{\mathrm{KL}}\!\left(q(\boldsymbol{\theta})\,\|\,p(\boldsymbol{\theta})\right).
\end{align*}

The resulting objective balances data-fit with the embedded constraints, alongside a regularization term on the parameters and constraint tolerance from the priors. The ELBO is minimized using the usual reparameterization trick \citep{kingmaAutoEncodingVariationalBayes2022}.

\paragraph{Interpretation.}
The framework may be interpreted as learning how much to trust prior knowledge. Rather than assuming perfect validity of constraints, the model learns the appropriate enforcement strength from the data. This is particularly important in real-world systems where constraints may hold only approximately due to measurement noise or partial observability. The resulting predictive distribution is both uncertainty-aware and constraint consistent in expectation.

\subsection{Uncertainty Decomposition Under Constraint Conditioning}
\label{sec:uncertainty-decomp}

Constraint conditioning modifies the structure of predictive uncertainty.
For a standard BNN, the law of total variance enables the following decomposition into aleatoric and epistemic terms:
\begin{align}
\mathrm{Var}(\mathbf{y}\mid\mathbf{x},\mathcal{D})
= \mathbb{E}_{q(\boldsymbol{\theta})}[\boldsymbol{\sigma}_P^2]
+ \mathrm{Var}_{q(\boldsymbol{\theta})}[\boldsymbol{\mu}_P],
\end{align}

Under constraint conditioning the predictive distribution becomes:
\begin{align}
p(\mathbf{y}\mid\mathbf{x},\mathcal{D})
= \mathbb{E}_{q(\boldsymbol{\theta}, \mathbf{r})}[\mathcal{N}(\boldsymbol{\mu}_C,\operatorname{diag}(\boldsymbol{\sigma}_C^2))],
\end{align}
with $\boldsymbol{\sigma}_C^2 = \boldsymbol{\sigma}_P^2 - \operatorname{diag}(\mathbf{K}\mathbf{S}\mathbf{K}^\top)$ following from Eqn.~\eqref{eq:sigma_update}.

Then, using the update rules in Eqns.~\eqref{eq:schur}--\eqref{eq:sigma_update} and applying the law of total variance over $q(\boldsymbol{\theta})q(\mathbf{r})$ gives:
\begin{align}
\mathrm{Var}(\mathbf{y})
= & \underbrace{\mathbb{E}_{q(\boldsymbol{\theta})}[\boldsymbol{\sigma}_P^2]}_{\text{Aleatoric}}
 - \underbrace{\mathbb{E}_{q(\boldsymbol{\theta}, \mathbf{r})}[\operatorname{diag}(\mathbf{K}\mathbf{S}\mathbf{K}^\top)]}_{\text{Constraint reduction}} \notag\\
& + \underbrace{\mathrm{Var}_{q(\boldsymbol{\theta})}[\boldsymbol{\mu}_P]}_{\text{Epistemic}} + \underbrace{\mathrm{Var}_{q(\mathbf{r})}[\boldsymbol{\mu}_C]}_{\text{Tolerance uncertainty}} \notag\\
& + \underbrace{\mathrm{Cov}_{q(\boldsymbol{\theta}, \mathbf{r})}[\boldsymbol{\mu}_P,\,\boldsymbol{\mu}_C-\boldsymbol{\mu}_P]}_{\text{Constraint--epistemic interaction}}.
\end{align}

Since each diagonal entry of $\mathbf{K}\mathbf{S}\mathbf{K}^\top$ is non-negative, constraint conditioning always reduces the marginal predictive variance of each output. As $\mathbf{r}\to 0$, the reduction is maximal (hard projection); and as $\mathbf{r}\to\infty$, it vanishes and the standard BNN decomposition is recovered.

\section{Experiments}

We evaluate our framework on learning the single particle model (SPM) of a lithium-ion battery, implemented via \texttt{PyBaMM} \citep{Sulzer2021}. The SPM is governed by spherical-diffusion equations, coupled with lumped thermal dynamics. The learning task approximates input-output mapping of the SPM: {\em Given operating conditions, predict electrochemical and thermal outputs subject to known physical constraints and noisy data}. We benchmark this against a standard unconstrained BNN.

\subsection{Problem Setup}

\paragraph{Inputs and outputs.}
We aim to learn the interaction between 3 input variables and 8 output variables within the SPM model, as summarised in Table~\ref{tab:io} in the Appendix. Data was generated by simulating full discharges across a grid of currents $C = \{0.5, 1.0, 1.5, 2.0, 2.5, 3.0\}$ and temperatures $T = \{273, 283, 293, 298, 303, 313, 318\}$, yielding $500$ state-of-charge points per combination. Of this, 60\% was used for training, 20\% for hyperparameter tuning and 20\% as reserved for a test set. Gaussian noise was added to the outputs to reflect realistic measurement uncertainty (voltage quantities $\sigma \approx 2$-$5$\,mV; thermal quantities $\sigma \approx 30$--$50$\,mW\,m$^{-3}$).

\paragraph{Constraints.} The SPM satisfies two linear equality constraints across the outputs, which we embed within the framework. Writing the output vector as $\mathbf{y} \in \mathbb{R}^8$, these take the form $\mathbf{B}\mathbf{y} = \mathbf{0}$ with $\mathbf{B} \in \mathbb{R}^{2 \times 8}$.

\textit{Kirchhoff's Voltage Law:}
\begin{align}\label{eq:Vctr}
V = V_{\mathrm{OCV}} - \eta_{+} - \eta_{-} - \Delta V_{\mathrm{IR}}.
\end{align}

\textit{Energy balance:}
\begin{align}\label{eq:Ectr}
\dot{Q}_{\mathrm{tot}} = \dot{Q}_{\mathrm{rev}} + \dot{Q}_{\mathrm{irr}}.
\end{align}

\paragraph{Chosen Priors and Variational Posteriors.}
We placed a standard Gaussian prior over the network parameters:
\[
p(\boldsymbol{\theta}) = \mathcal{N}(\mathbf{0}, \mathbf{I}_P),
\]
where $P$ denotes the number of parameters.

To ensure positivity of the constraint tolerance, we reparameterized
$\mathbf{r}$ via a log-scale variable $\boldsymbol{\rho}$ such that
\[
\mathbf{r} = \exp(\rho).
\]
We also placed a Gaussian prior over $\boldsymbol{\rho}$:
\[
p(\rho) = \mathcal{N}(\boldsymbol{\mu}_{\boldsymbol{\rho}}, \boldsymbol{\Sigma}_{\boldsymbol{\rho}}),
\]
with $\boldsymbol{\mu}_{\boldsymbol{\rho}} = [-2, -2]^\top$ and
$\boldsymbol{\sigma}_{\boldsymbol{\rho}} = \operatorname{diag}([1,1])$,
reflecting a prior belief that the constraints hold approximately, with limited dispersion around the mean.

Both variational posteriors $q(\boldsymbol{\theta})$ and $q(\boldsymbol{\rho})$ were chosen as mean-field diagonal Gaussians.

\subsection{Results}

We compare all results on the test set. The standard BNN and the proposed Bayesian constrained probabilistic neural network (BCPNN) are comparable in terms of point predictive accuracy, and they achieve identical performance on coverage ratio (0.99), indicating that embedding constraints does not degrade regression performance. 

When evaluating the predictive distributions (Table~\ref{tab:prediction}), the BCPNN also produces tighter coverage widths, reflecting the variance reduction predicted by the decomposition in Section~\ref{sec:uncertainty-decomp}. When sampling over the variational posteriors, we also see the BCPNN reduces both aleatoric uncertainty due to Gaussian conditioning on the constraint, and epistemic uncertainty, reflecting tighter posterior concentration and increased predictive confidence. 

\begin{table}[htb]
\centering
\footnotesize
\caption{\label{tab:prediction}Predictive uncertainty decomposition, in normalized space.}
\begin{tabular}{lcc}
\toprule
\textbf{Model} & \textbf{Aleatoric} & \textbf{Epistemic} \\
\midrule
BNN & $0.0149 \pm 0.0096$ & $0.0074 \pm 0.0042$ \\
BCPNN & $\mathbf{0.0140} \pm \mathbf{0.0097}$ & $\mathbf{0.0068} \pm \mathbf{0.0041}$ \\
\bottomrule
\end{tabular}
\end{table}

However, the most significant distinction between these models lies in constraint adherence. The BNN exhibits large violations on the test set, as it has no mechanism to enforce the voltage decomposition or heat balance beyond what is implicitly learned from data. The BCPNN reduces median violation by over two orders of magnitude, for the voltage constraint, and over four on the heat balance (Table~\ref{tab:constraint_violation}), demonstrating that probabilistic conditioning enforces clear physical consistency. 

The learned tolerance posteriors (Table~\ref{tab:learned_tolerance}) further show that the framework is able to automatically distinguish between constraints: Kirchhoff's voltage law (Eqn.~\ref{eq:Vctr}) is enforced to a higher precision, while the energy balance (Eqn.~\ref{eq:Ectr}) retains appreciable slack with high posterior uncertainty. This result is expected due to differing measurement noise across variables, illustrating the framework's ability to calibrate constraint confidence directly from the data.

\begin{table}[htbp]
\centering
\footnotesize
\caption{\label{tab:learned_tolerance}Learned constraint tolerance posteriors.}
\begin{tabular}{lcccc}
\toprule
& $\boldsymbol{\mu_\rho}$ & $\boldsymbol{\sigma_\rho}$ & $\boldsymbol{\mathbb{E}[\mathbf{r}]}$ & $\textbf{Std}[\mathbf{r}]$ \\
\midrule
Eqn.~\eqref{eq:Vctr} & $-11.21$ & $0.23$ & $1.36 \times 10^{-5}$ & $3.19 \times 10^{-6}$ \\
Eqn.~\eqref{eq:Ectr} & $-1.99$  & $1.00$ & $1.37 \times 10^{-1}$ & $1.36 \times 10^{-1}$ \\
\bottomrule
\end{tabular}
\end{table}

\begin{figure}[htb]
    \centering
    \includegraphics[width=1\linewidth]{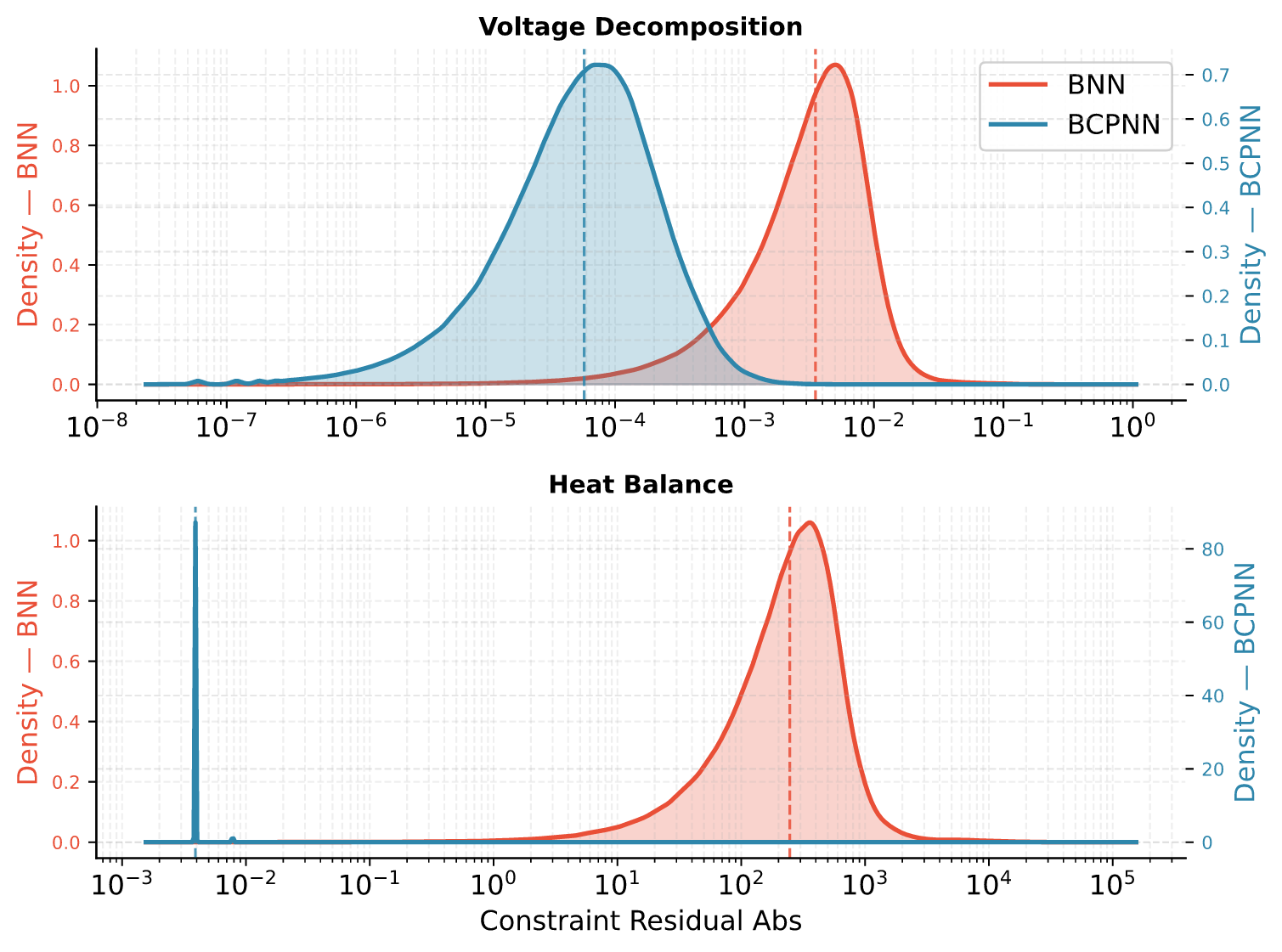}
    \caption{KDE over sampled posteriors of constraint violations}
    \label{fig:placeholder}
\end{figure}

\section{Conclusions}

We presented a framework for incorporating linear equality constraints into BNNs via Gaussian conditioning. The resulting BCPNN preserves the predictive accuracy of a standard BNN while reducing constraint violations and producing much tighter and physically consistent uncertainty estimates. We also showed how the learned tolerance posteriors automatically can distinguish between constraints of different tolerances, providing interpretable diagnostics of constraint confidence without manual tuning.

The current formulation is applied to only linear equality constraints here. Extending the conditioning framework to handle inequality constraints and nonlinear relationships would broaden its applicability. Additionally, integrating the learned tolerance posteriors into active learning or experimental design pipelines could enable data collection strategies that are informed by both predictive, and constraint uncertainty could provide interesting avenues for further exploration.

\subsubsection*{Acknowledgements}
Thanks to Dr. Laura Helleckes for her help reviewing the manuscript.

\footnotesize
\bibliographystyle{plainnat}
\bibliography{references}

@InProceedings{blundellWeightUncertaintyNeural,
  title = 	 {Weight Uncertainty in Neural Network},
  author = 	 {Blundell, Charles and Cornebise, Julien and Kavukcuoglu, Koray and Wierstra, Daan},
  booktitle = 	 {Proceedings of the 32nd International Conference on Machine Learning},
  pages = 	 {1613--1622},
  year = 	 {2015},
  editor = 	 {Bach, Francis and Blei, David},
  volume = 	 {37},
  optseries = 	 {Proceedings of Machine Learning Research},
  address = 	 {Lille, France},
  month = 	 {07--09 Jul},
  optpublisher =    {PMLR},
  opturl = 	 {https://proceedings.mlr.press/v37/blundell15.html}
}

@misc{paszkePyTorchImperativeStyle2019,
  title = {{{PyTorch}}: {{An Imperative Style}}, {{High-Performance Deep Learning Library}}},
  shorttitle = {{{PyTorch}}},
  author = {Paszke, Adam and Gross, Sam and Massa, Francisco and Lerer, Adam and Bradbury, James and Chanan, Gregory and Killeen, Trevor and Lin, Zeming and Gimelshein, Natalia and Antiga, Luca and Desmaison, Alban and K{\"o}pf, Andreas and Yang, Edward and DeVito, Zach and Raison, Martin and Tejani, Alykhan and Chilamkurthy, Sasank and Steiner, Benoit and Fang, Lu and Bai, Junjie and Chintala, Soumith},
  year = 2019,
  month = dec,
  number = {arXiv:1912.01703},
  eprint = {1912.01703},
  primaryclass = {cs},
  publisher = {arXiv},
  doi = {10.48550/arXiv.1912.01703},
  urldate = {2025-05-08},
  archiveprefix = {arXiv}
}

@book{nealBayesianLearningNeural1996,
  title = {Bayesian {{Learning}} for {{Neural Networks}}},
  author = {Neal, Radford M.},
  editor = {Bickel, P. and Diggle, P. and Fienberg, S. and Krickeberg, K. and Olkin, I. and Wermuth, N. and Zeger, S.},
  year = 1996,
  series = {Lecture {{Notes}} in {{Statistics}}},
  volume = {118},
  publisher = {Springer},
  address = {New York, NY},
  doi = {10.1007/978-1-4612-0745-0},
  opturldate = {2026-02-24},
  optcopyright = {http://www.springer.com/tdm},
  optisbn = {978-0-387-94724-2 978-1-4612-0745-0}
}

@article{Sulzer2021,
  title = {{Python battery mathematical modelling (PyBaMM)}},
  author = {Sulzer, Valentin and Marquis, Scott G. and Timms, Robert and Robinson, Martin and Chapman, S. Jon},
  doi = {10.5334/jors.309},
  journal = {Journal of Open Research Software},
  publisher = {Software Sustainability Institute},
  volume = {9},
  number = {1},
  pages = {14},
  year = {2021}
}

@article{raissiPhysicsinformedNeuralNetworks2019,
title = {Physics-informed neural networks: {A} deep learning framework for solving forward and inverse problems involving nonlinear partial differential equations},
volume = {378},
optissn = {0021-9991},
optshorttitle = {Physics-informed neural networks},
opturl = {https://www.sciencedirect.com/science/article/pii/S0021999118307125},
doi = {10.1016/j.jcp.2018.10.045},
journal = {Journal of Computational Physics},
author = {Raissi, M. and Perdikaris, P. and Karniadakis, G. E.},
year = {2019},
pages = {686--707},
}

@inproceedings{amosOptNetDifferentiableOptimization2017,
title = {{OptNet}: {Differentiable} {Optimization} as a {Layer} in {Neural} {Networks}},
optissn = {2640-3498},
optshorttitle = {{OptNet}},
opturl = {https://proceedings.mlr.press/v70/amos17a.html},
language = {en},
opturldate = {2025-04-25},
booktitle = {Proceedings of the 34th {International} {Conference} on {Machine} {Learning}},
opt = {PMLR},
author = {Amos, Brandon and Kolter, J. Zico},
month = jul,
year = {2017},
pages = {136--145},
}

@misc{dontiDC3LearningMethod2021,
	title = {{DC3}: {A} learning method for optimization with hard constraints},
	optshorttitle = {{DC3}},
	opturl = {http://arxiv.org/abs/2104.12225},
	doi = {10.48550/arXiv.2104.12225},
	opturldate = {2025-04-08},
	publisher = {arXiv},
	author = {Donti, Priya L. and Rolnick, David and Kolter, J. Zico},
	month = apr,
	year = {2021},
	note = {arXiv:2104.12225 [cs]},
}

@inproceedings{gravesPracticalVariationalInference2011,
doi = {10.5555/2986459.2986721},
author = {Graves, Alex},
title = {Practical variational inference for neural networks},
year = {2011},
optisbn = {9781618395993},
optpublisher = {Curran Associates Inc.},
address = {Red Hook, NY, USA},
booktitle = {Proceedings of the 25th International Conference on Neural Information Processing Systems},
pages = {2348–2356},
optnumpages = {9},
optlocation = {Granada, Spain},
optseries = {NIPS'11}
}

@inproceedings{galDropoutBayesianApproximation2016,
	title = {Dropout as a {Bayesian} Approximation: {Representing} Model Uncertainty in Deep Learning},
	optissn = {1938-7228},
	optshorttitle = {Dropout as a {Bayesian} {Approximation}},
	opturl = {https://proceedings.mlr.press/v48/gal16.html},
	optlanguage = {en},
	opturldate = {2025-06-06},
	booktitle = {Proceedings of The 33rd {International} {Conference} on {Machine} {Learning}},
	publisher = {PMLR},
	author = {Gal, Yarin and Ghahramani, Zoubin},
	month = jun,
	year = {2016},
	pages = {1050--1059},
}

@incollection{agrawalDifferentiableConvexOptimization2019,
	address = {Red Hook, NY, USA},
	title = {Differentiable convex optimization layers},
	number = {858},
	urldate = {2025-04-25},
	booktitle = {Proceedings of the 33rd {International} {Conference} on {Neural} {Information} {Processing} {Systems}},
	publisher = {Curran Associates Inc.},
	author = {Agrawal, Akshay and Amos, Brandon and Barratt, Shane and Boyd, Stephen and Diamond, Steven and Kolter, J. Zico},
	month = dec,
	year = {2019},
	pages = {9562--9574},
}

@article{chenPhysicsinformedNeuralNetworks2024,
	title = {Physics-informed neural networks with hard linear equality constraints},
	volume = {189},
	optissn = {0098-1354},
	opturl = {https://www.sciencedirect.com/science/article/pii/S0098135424001820},
	doi = {10.1016/j.compchemeng.2024.108764},
	urldate = {2025-04-25},
	journal = {Computers \& Chemical Engineering},
	author = {Chen, Hao and Flores, Gonzalo E. Constante and Li, Can},
	month = oct,
	year = {2024},
	pages = {108764},
}

@inproceedings{hansenLearningPhysicalModels2023,
	title = {Learning Physical Models that Can Respect Conservation Laws},
	optissn = {2640-3498},
	opturl = {https://proceedings.mlr.press/v202/hansen23b.html},
	language = {en},
	urldate = {2025-07-23},
	booktitle = {Proceedings of the 40th {International} {Conference} on {Machine} {Learning}},
	optpublisher = {PMLR},
	author = {Hansen, Derek and Maddix, Danielle C. and Alizadeh, Shima and Gupta, Gaurav and Mahoney, Michael W.},
	month = jul,
	year = {2023},
	pages = {12469--12510},
}

@book{pml2Book,
	title = {Probabilistic Machine Learning: {Advanced} Topics},
	opturl = {http://probml.github.io/book2},
	publisher = {MIT Press},
	author = {Murphy, Kevin P.},
	year = {2023},
}

@misc{kingmaAutoEncodingVariationalBayes2022,
	title = {Auto-Encoding Variational Bayes},
	opturl = {http://arxiv.org/abs/1312.6114},
	doi = {10.48550/arXiv.1312.6114},
	opturldate = {2025-06-10},
	optpublisher = {arXiv},
	author = {Kingma, Diederik P. and Welling, Max},
	month = dec,
	year = {2022},
	note = {arXiv:1312.6114 [stat]},
}

\clearpage
\appendix
\thispagestyle{empty}

\onecolumn
\aistatstitle{Learning with Embedded Linear Equality Constraints via Variational Bayesian Inference: \\
Supplementary Materials}

\section{Derivations of Results}

\subsection{Derivation of Modified ELBO}

We seek a tractable approximation to the true posterior, as the minimizer of the KL-divergence:
\begin{align*}
q^*(\boldsymbol{\theta}, \mathbf{r})
&\in \arg\min_{q(\boldsymbol{\theta}, \mathbf{r})}
D_{\mathrm{KL}}\!\left(
q(\boldsymbol{\theta}, \mathbf{r})\,\|\, p(\boldsymbol{\theta}, \mathbf{r} \mid \mathcal{D})
\right)\\ & =
\int \int
q(\boldsymbol{\theta}, \mathbf{r})
\log
\frac{q(\boldsymbol{\theta}, \mathbf{r})}
     {p(\boldsymbol{\theta}, \mathbf{r} \mid \mathcal{D})}
\, d\boldsymbol{\theta} \, d\mathbf{r},
\end{align*}
using Bayes' rule $p(\boldsymbol{\theta}, \mathbf{r} \mid \mathcal{D})
\propto
p(\mathcal{D} \mid \boldsymbol{\theta}, \mathbf{r})\, p(\boldsymbol{\theta}, \mathbf{r})$, where $p(\mathcal{D} \mid \boldsymbol{\theta}, \mathbf{r})$ is our augmented likelihood. 

Substituting into the KL objective gives:
\begin{align*}
\int \int
q(\boldsymbol{\theta}, \mathbf{r})
\log
\frac{q(\boldsymbol{\theta}, \mathbf{r})}
     {p(\mathcal{D} \mid \boldsymbol{\theta}, \mathbf{r})\, p(\boldsymbol{\theta}, \mathbf{r})}
\, d\boldsymbol{\theta} \, d\mathbf{r}
&=
\mathbb{E}_{q(\boldsymbol{\theta}, \mathbf{r})}
\left[
\log
\frac{q(\boldsymbol{\theta}, \mathbf{r})}
     {p(\mathcal{D} \mid \boldsymbol{\theta}, \mathbf{r})\, p(\boldsymbol{\theta}, \mathbf{r})}
\right]
\\
&=
\mathbb{E}_{q(\boldsymbol{\theta}, \mathbf{r})}
\left[
\log
\frac{q(\boldsymbol{\theta}, \mathbf{r})}
     {p(\boldsymbol{\theta}, \mathbf{r})}
\right] - \mathbb{E}_{q(\boldsymbol{\theta}, \mathbf{r})}
\left[
\log p(\mathcal{D} \mid \boldsymbol{\theta}, \mathbf{r})\
\right].
\end{align*}

Then, using a mean-field factorization
$q(\boldsymbol{\theta}, \mathbf{r})=q(\boldsymbol{\theta})q(\mathbf{r})$ and
$p(\boldsymbol{\theta}, \mathbf{r})=p(\boldsymbol{\theta})p(\mathbf{r})$, we obtain:
\begin{align*}
\int \int
q(\boldsymbol{\theta}, \mathbf{r})
\log
\frac{q(\boldsymbol{\theta}, \mathbf{r})}
     {p(\mathcal{D} \mid \boldsymbol{\theta}, \mathbf{r})\, p(\boldsymbol{\theta}, \mathbf{r})}
\, d\boldsymbol{\theta} \, d\mathbf{r}
&=
\mathbb{E}_{q(\boldsymbol{\theta})q(\mathbf{r})}
\left[
\log
\frac{q(\boldsymbol{\theta})q(\mathbf{r})}
     {p(\boldsymbol{\theta})p(\mathbf{r})}
\right] - \mathbb{E}_{q(\boldsymbol{\theta})q(\mathbf{r})}
\left[
\log p(\mathcal{D} \mid \boldsymbol{\theta}, \mathbf{r})\
\right]
\\
&=
\mathbb{E}_{q(\boldsymbol{\theta})q(\mathbf{r})}
\!\left[
-\log p(\mathcal{D} \mid \boldsymbol{\theta}, \mathbf{r})
\right] + D_{\mathrm{KL}}\!\left(q(\boldsymbol{\theta})\,\|\,p(\boldsymbol{\theta})\right)
+ D_{\mathrm{KL}}\!\left(q(\mathbf{r})\,\|\,p(\mathbf{r})\right).
\end{align*}

\subsection{Derivation of the Uncertainty Decomposition}

The predictive distribution of a new output $\mathbf{y}^*$ given input
$\mathbf{x}^*$ is
\begin{align}
p(\mathbf{y}^* \mid \mathbf{x}^*, \mathcal{D})
&=
\int \int
p(\mathbf{y}^* \mid \mathbf{x}^*, \boldsymbol{\theta}, \mathbf{r})
\, q(\boldsymbol{\theta}) q(\mathbf{r})
\, d\boldsymbol{\theta} \, d\mathbf{r} .
\end{align}

Since for fixed $(\boldsymbol{\theta}, \mathbf{r})$ the predictive distribution is Gaussian:
\begin{equation}
p(\mathbf{y}^* \mid \mathbf{x}^*, \boldsymbol{\theta}, \mathbf{r})
=
\mathcal{N}\!\left(
\boldsymbol{\mu}_C,
\boldsymbol{\sigma}^2_C
\right),
\end{equation}
the law of total variance gives
\begin{align}
\operatorname{Var}(\mathbf{y}^* \mid \mathbf{x}^*, \boldsymbol{\theta}, \mathbf{r})
&=
\mathbb{E}_{q(\boldsymbol{\theta})q(\mathbf{r})}
\!\left[
\boldsymbol{\sigma}^2_C
\right]
+
\operatorname{Var}_{q(\boldsymbol{\theta})q(\mathbf{r})}
\!\left[
\boldsymbol{\mu}_C
\right].
\label{eq:total_var}
\end{align}

Then, from Gaussian conditioning, we have:
\begin{align}
\mathbf{S}
&=
\mathbf{B}
\operatorname{diag}(\boldsymbol{\sigma}_P^2)
\mathbf{B}^{\top}
+
R,
\\
\boldsymbol{\mu}_C
&=
\boldsymbol{\mu}_P
+
\mathbf{K}
\left(
\mathbf{b}
-
\mathbf{A}\mathbf{x}^*
-
\mathbf{B}\boldsymbol{\mu}_P
\right),
\\
\boldsymbol{\sigma}_C^2
&=
\operatorname{diag}(\boldsymbol{\sigma}_P^2)
-
\mathbf{K}
\mathbf{S}
\mathbf{K}^{\top},
\end{align}
where
\begin{equation}
\mathbf{K}
=
\operatorname{diag}(\boldsymbol{\sigma}_P^2)
\mathbf{B}^{\top}
\mathbf{S}^{-1}.
\end{equation}

Since we only consider the diagonal elements of the covariance matrix, substituting $\boldsymbol{\sigma}^2_C$ into Eqn.~\eqref{eq:total_var} yields
\begin{align}
\operatorname{Var}(\mathbf{y}^* \mid \mathbf{x}^*, \boldsymbol{\theta}, \mathbf{r})
&=
\mathbb{E}_{q(\boldsymbol{\theta})q(\mathbf{r})}
\!\left[
\operatorname{diag}(\boldsymbol{\sigma}_P^2)
\right]
-
\mathbb{E}_{q(\boldsymbol{\theta})q(\mathbf{r})}
\!\left[
\mathbf{K}
\mathbf{S}
\mathbf{K}^{\top}
\right]
+
\operatorname{Var}_{q(\boldsymbol{\theta})q(\mathbf{r})}
\!\left[
\boldsymbol{\mu}_C
\right].
\label{eq:variance_expanded}
\end{align}

The first expectation corresponds to aleatoric uncertainty inherited
from the network likelihood, while the second term is a positive
semi-definite reduction arising from constraint conditioning.

We now expand the mean-variance term. Writing
\begin{equation}
\boldsymbol{\mu}_C
=
\boldsymbol{\mu}_P
+
\Delta(\boldsymbol{\theta}, \mathbf{r}),
\end{equation}
where
\begin{equation}
\Delta(\boldsymbol{\theta}, \mathbf{r})
=
\mathbf{K}
\left(
\mathbf{b}
-
\mathbf{A}\mathbf{x}^*
-
\mathbf{B}\boldsymbol{\mu}_P
\right),
\end{equation}
we obtain
\begin{align}
\operatorname{Var}_{q(\boldsymbol{\theta})q(\mathbf{r})}(\boldsymbol{\mu}_C)
&=
\operatorname{Var}_{q(\boldsymbol{\theta})}(\boldsymbol{\mu}_P)
+
\operatorname{Var}_{q(\boldsymbol{\theta})q(\mathbf{r})}(\Delta)
+
2\,\operatorname{Cov}_{q(\boldsymbol{\theta})q(\mathbf{r})}
\big(
\boldsymbol{\mu}_P,
\Delta
\big).
\end{align}

Combining terms, the predictive variance decomposes as
\begin{align}
\operatorname{Var}(\mathbf{y}^*)
&=
\underbrace{
\mathbb{E}_{q(\boldsymbol{\theta})}
\!\left[
\operatorname{diag}(\boldsymbol{\sigma}_P^2)
\right]
}_{\text{Aleatoric}}
+
\underbrace{
\operatorname{Var}_{q(\boldsymbol{\theta})}
(\boldsymbol{\mu}_P)
}_{\text{Epistemic}}
-
\underbrace{
\mathbb{E}_{q(\boldsymbol{\theta})q(\mathbf{r})}
\!\left[
\mathbf{K}
\mathbf{S}
\mathbf{K}^{\top}
\right]
}_{\text{Constraint Reduction}}
+
\underbrace{
\operatorname{Var}_{q(\boldsymbol{\theta})q(\mathbf{r})}(\Delta)
}_{\text{Constraint Tolerance}}
+
\underbrace{
2\,\operatorname{Cov}_{q(\boldsymbol{\theta})q(\mathbf{r})}
(\boldsymbol{\mu}_P,\Delta)
}_{\text{Interaction}}.
\end{align}

Since $\mathbf{K}\mathbf{S}\mathbf{K}^{\top}$ is positive semi-definite,
constraint conditioning always reduces predictive uncertainty in
directions aligned with the constraint. The remaining terms quantify
uncertainty induced by posterior variability in the constraint tolerance
and its interaction with parameter uncertainty.

\section{Numerical Experiments}

\subsection{Computational Setup}
All experiments were carried out on a PC, equipped with 32GB DDR5 RAM, with an Intel Core i9 CPU and an NVIDIA RTX5090 GPU with 32GB of DDR6 VRAM. All models were implemented in \texttt{PyTorch} \cite{paszkePyTorchImperativeStyle2019}

\subsection{Input and Output Variables}

The variables used within the SPM model, and their applicability to our learning framework, are given in Table~\ref{tab:io}. Simulations were performed under heteroscedastic observation noise applied independently to each output variable.

\begin{table}[htb]
\centering
\footnotesize
\caption{SPM inputs and outputs used in the learning.}
\label{tab:io}
\begin{tabular}{lllll}
\toprule
\textbf{} & \textbf{Symbol} & \textbf{Description} & \textbf{Unit} & \textbf{Range} \\
\midrule
\multicolumn{4}{l}{\textbf{Inputs}} \\
& $I$   & Applied current       & A  & $0.5$--$3.0$ \\
& SOC   & State of charge       & -- & $0.05$--$0.95$ \\
& $T$   & Ambient temperature   & K  & $273$--$318$ \\
\multicolumn{4}{l}{\textbf{Outputs}} \\
& $V$                           & Terminal voltage                 & V \\
& $V_{\mathrm{OCV}}$            & Open-circuit voltage             & V \\
& $\eta_{+}$                    & Positive electrode overpotential & V \\
& $\eta_{-}$                    & Negative electrode overpotential & V \\
& $\Delta V_{\mathrm{IR}}$      & Ohmic (IR) drop                  & V \\
& $\dot{Q}_{\mathrm{tot}}$      & Total volumetric heating         & W\,m$^{-3}$ \\
& $\dot{Q}_{\mathrm{rev}}$      & Reversible heating               & W\,m$^{-3}$ \\
& $\dot{Q}_{\mathrm{irr}}$      & Irreversible heating             & W\,m$^{-3}$ \\
\bottomrule
\end{tabular}
\end{table}

\subsection{Predictive Performance}

We evaluate predictive performance across all outputs using mean squared error (MSE) and credible interval width (CW). Results are reported as mean $\pm$ $1.96\sigma$ across output variables and are shown in Table~\ref{tab:predictive_performance}.
\begin{table}[h]
\centering
\footnotesize
\caption{Model predictive performance ($\mu \pm 1.96\sigma$ over outputs).}
\label{tab:predictive_performance}
\begin{tabular}{lcc}
\toprule
\textbf{Model} & \textbf{MSE} & \textbf{CW} \\
\midrule
BNN    & $1.13 \times 10^{-4} \pm 1.88 \times 10^{-3}$ & $0.067 \pm 0.022$ \\
BCPNN  & $\mathbf{1.11 \times 10^{-4} \pm 2.21 \times 10^{-3}}$ & $\mathbf{0.059 \pm 0.022}$ \\
\bottomrule
\end{tabular}
\end{table}

\subsection{Constraint Satisfaction}

To assess structural consistency, we sampled 10,000 Monte Carlo posterior draws and computed the corresponding constraint residuals across the test set. The mean violation magnitude is reported in Table~\ref{tab:constraint_violation}.
\begin{table}[b]
\centering
\footnotesize
\label{tab:constraint_violation}
\begin{tabular}{lc}
\toprule
\textbf{Model} & \textbf{Constraint Violation} \\
\midrule
BNN    & $174.82 \pm 90.78$ \\
BCPNN  & $\mathbf{8.81 \times 10^{-4} \pm 4.3 \times 10^{-5}}$ \\
\bottomrule
\end{tabular}
\end{table}
\end{document}